\documentclass{article}

\usepackage{algorithm}

\usepackage[final]{nips_2016}
\usepackage[utf8]{inputenc} 
\usepackage[T1]{fontenc}    
\usepackage{hyperref}       
\usepackage{url}            
\usepackage{booktabs}       
\usepackage{amsfonts}       
\usepackage{nicefrac}       
\usepackage{microtype}      
\usepackage{xcolor}
\usepackage{natbib}

\title{Reinforcement Learning for Transition-Based Mention Detection}

\author{
  Georgiana Dinu, Wael Hamza and Radu Florian \\
  IBM T.J. Watson Research Center\\
  1101 Kitchawan Rd, Yorktown Heights, NY 10598\\
  \texttt{whamza|gdinu|raduf@us.ibm.com} \\
  }

\begin{document}

\maketitle

\begin{abstract}
This paper describes an application of reinforcement learning to the mention detection task. We define a novel action-based formulation for the mention detection task, in which a model can flexibly revise past labeling decisions by grouping together tokens and assigning partial mention labels. We devise a method to create mention-level episodes and we train a model by rewarding correctly labeled complete mentions, irrespective of the inner structure created. The model yields results which are on par with a competitive supervised counterpart while being more flexible in terms of achieving targeted behavior through reward modeling and generating internal mention structure, especially on longer mentions. 
\end{abstract}

\section{Introduction}
Mention detection is the natural language processing task of identifying and assigning labels to occurrences of entities in text, according to a predefined type-system. For example the following sentence (from \citet{ontonotes}): \textit{Monday night, Colin Powel, Robert Downey Junior and Sharon Stone.} contains three mentions of type PERSON: Colin Powel, Robert Downey Junior, Sharon Stone and a mention labeled TIME: Monday night. As this is a joint chunking and labeling task, the standard approach is to formulate it as a sequence labeling problem, in which the label of a mention word encodes both the type and the position within the mention. In the previous example, a correct PER classification within the BIO encoding scheme is Robert$_{\mathrm{B-PER}}$ Downey$_{\mathrm{I-PER}}$ Junior$_{\mathrm{I-PER}}$.

While performance for this task is relatively high for short mentions, longer mentions are considerably more difficult to label correctly. Consider for example the sentence: \textit{The George Washington Bridge is owned and operated by the New York and New Jersey Port Authority.} A standard annotation for this sentence labels [George Washington Bridge]$_{\mathrm{FACILITY}}$ and [New York and New Jersey Port Authority]$_{\mathrm{ORG}}$ as mentions. Such examples pose difficulties to models due to the inherent bias created by, for example, New York mostly occurring as GPE. This is still a problem within neural network architectures (\citet{andor2016globally}). In order to mitigate the label bias problems (\citet{bottou91,Lafferty:2001:CRF}), models may require significant look-ahead and increased capacity (to encode that e.g. New is an ORG when followed by Port Authority or that George in the context of Bridge is a FACILITY) or global normalization with e.g. CRF objectives.

Alternatively, having the capacity to generate internal structure and update labels as this structure is built may allow a model to more naturally revise labeling decisions. For example, a correct label can be obtained by initially labeling George Washington as a PERSON and then composing [PERSON Bridge] into FACILITY, both decisions corresponding to frequent enough patterns which a model should be able to learn more easily from training data. For this purpose, we propose a transition-based formulation which can incrementally construct mentions as labeled binary trees and which is trained in a reinforcement learning setting. Specifically, the labeling is obtained as the output of a sequence of actions, allowing it to either label an incoming token or compose the 2 nodes on the top of the current stack and assign it a (potentially different) label. As inner structure is not usually annotated, and hence cannot be learned directly, a reinforcement learning setting is the natural choice for learning and predicting it. In this paper, we propose a Q-function approximating network which positively rewards a sequence of actions which predicts a correct mention tree root. Crucially, we do not assign any reward to partial decisions in order to encourage the model to freely learn to label incomplete mention spans. 

The remainder of the paper is organized as follows: Section 2 discusses relevant literature. Section 3 introduces our proposal for mention detection through reinforcement learning while Sections 4 and 5 present the experimental evaluation and conclude the paper.


\section{Related work}
Mention detection and named entity recognition\footnote{The main difference between the two is that mention detection has a larger scope: besides named mentions, it also identifies nominal and pronominal mentions.}, have been extensively studied in the literature, due to their usefulness to end-user tasks (such as question answering, converting unstructured information into structured data, etc). Datasets for named entity recognition have been created since the mid 90s, with the MUC-6 dataset, then continued with the CoNLL datasets (\cite{conll02,conll03}), the ACE dataset (\cite{ACE-2003}), and the OntoNotes dataset (\cite{ontonotes}). The problem has also been studied across languages (English, Arabic, Chinese, Spanish, German, Dutch, and Hindi are among the most studied languages). Most of the approaches treat the task as a sequence labeling task, where each word gets assigned a tag corresponding to whether it is a non-mention word (usually, "O"), starts a mention or continues a mention. This corresponds to a reversible mapping from the original chunks to word-based labels (this BIO scheme was first described by \cite{ramshaw94} for noun phrase chunking), and leads to standard sequence-based classification models such as maximum entropy \citet{borthwick98exploiting,ibmace04}, CRFs (\citep{lafferty01:_condit}) and neural-network based approaches (\citep{Collobert:2011:NLP:1953048.2078186}). Approaches that attempt to produce the chunks directly have also been investigated, such as the one by \cite{Daume:2005:LEE:1220575.1220588}.


A transition-based model for NER as an alternative to the standard encoding has been  introduced in \citet{Lample16}, where it is shown to obtained state-of-the-art results among comparable models and it is only outperformed by a globally normalized LSTM model. The formulation is inspired by dependency parsing in which the labeling is the result of a sequence of \textit{transitions} which at each step decide whether to mark an incoming word as non-mention, start/continue a mention or end a mention and assign a label. Under this formulation, for each labeled sequence there is an unique sequence of transitions which generates the correct annotation and therefore it is natural to remain within a supervised learning setting.


Reinforcement learning for sequence labeling problems in NLP (including Named Entity Recognition) has been proposed in \citet{Maes09}. Unlike our proposal, the action space closely follows the supervised NER formulation consisting of token-level labels using the BIO encoding, assigned either in a left-to-right order or in a free-order fashion. Rewards are given for each episode or for each action (positive if the label assigned is the same as the gold standard). They obtain competitive results when using left-to-right order and action-level rewards which outperform the more flexible free-order and episode reward alternatives.

Our work comes closer to that of \citet{Jiarong12} in which the authors propose a hybrid reinforcement/apprenticeship learning algorithm for transition-based dependency parsing in order to obtain a better speed-accuracy trade-off. As a policy gradient algorithm as well as an apprenticeship learning approach fail to reach competitive performance, the authors propose an oracle-infused hybrid which during exploration uses the oracle to generate the correct action with fixed probability.  



\section{Mention Detection and Reinforcement Learning}
We formulate the problem as a Markov Decision Process (MDP) $\left(\mathcal{S},\mathcal{A},\mathcal{R}\right)$, where $\mathcal{S}$ is the state space with discrete actions $\mathcal{A}$ and a reward function $\mathcal{R}:\mathcal{A}\times\mathcal{S}\to \mathbb{R}$. We learn the state-action value function $Q$ and our policy $\pi$ is a greedy policy in $Q$. The following section describes the formulation of the transition-based mention detection problem as an MDP.
\subsection{MDPs for mention detection}
\paragraph{Transition Based Mention Detection} The mention detection agent scans the given sentence from beginning to end making labeling decisions, given by the action space, sequentially and one word at a time. During learning, the agent is given rewards based on the gold annotation in the training data. The decision process terminates when the process consumes the entire sentence. 
\paragraph{States} We represent the current state using a stack and a buffer (using the terminology in \citet{Lample16}). The stack contains the sequence of tokens which have already been labeled, together with their labels, while the buffer consists of the words left to be processed. Table \ref{action-example} contains an example of a buffer and stack values for a transition sequence example. The neural network approximating the $Q$ function uses features extracted from the current state, typically some features of the current word (top of the buffer) and several past tokens (the top elements of the stack). The specific input representation used is detailed in Section \ref{exp-details}.
\paragraph{Action space} Given a label set $\mathcal{L}$ we define an action space of size 2x$|\mathcal{L}|$: 
$$ \mathcal{A} = \{\mathrm{Shift}_{\mathrm{LABEL}}, \mathrm{Reduce}_{\mathrm{LABEL}}| \mathrm{LABEL}\in \mathcal{L}\}$$
Shift actions move a token from the top of the buffer to the stack and assign it the class LABEL. Reduce actions group together and label the top 2 elements of the stack, which can be multi-word sequences themselves. There is no restriction on the $\mathrm{Reduce}_\mathrm{LABEL}$ to be consistent with the classes already assigned to the stack elements, i.e. the model can decide to re-label elements - for instance, joining a $\mathrm{PERSON}$ group with an $\mathrm{O}$ group to form a $\mathrm{FACILITY}$ group.
A sequence of actions is uniquely mapped to a label sequence and it is read off the stack at the end of an episode. Table \ref{action-example} contains an example. Non-mention (O labels) are treated like any other labels with the exception of Reduce-O actions which are not allowed. 
\begin{table}[h]
\caption{Transition sequence example for \textit{George Washington Bridge}}
  \label{action-example}
  \centering
  \begin{tabular}{lll}
    \toprule
    Stack & Buffer & Decision  \\
    \midrule
$\emptyset$ & George Washington Bridge & Shift-PER\\
{George}$_{\mathrm{PER}}$ & Washington Bridge & Shift-PER\\
{George}$_{\mathrm{PER}}$ {Washington}$_{\mathrm{PER}}$  & Bridge & Reduce-PER\\
{$[$George Washington$]$}$_{\mathrm{PER}}$ & Bridge & Shift-O\\
{$[$George Washington$]$}$_{\mathrm{PER}}$ {Bridge}$_{\mathrm{O}}$ & $\emptyset$ & Reduce-FAC\\
{$[$George Washington Bridge$]$}$_{\mathrm{FAC}}$& $\emptyset$ & -\\
    \bottomrule
  \end{tabular}
\end{table}
\paragraph{Episodes and rewards}
At test time, the episode is defined on the sentence level, that is, the agent starts at the beginning of the sentence and continues to make decisions until the entire sentence is consumed. However, during learning, we explore two types of episodes: \textit{sentence} episodes (used at test time) and \textit{mention} episodes. For sentence episodes the reward can simply be the F-score of the sentence (or a function thereof). For mention-level episodes, the agent starts the episode from the beginning of a mention (not necessarily at the beginning of the sentence), and we assign positive rewards if the entire mention is correctly labeled, negative otherwise. We do not assign any reward for intermediate decisions within the mention boundaries. In one of our experiments, however, we experimented with assigning partial reward in cases where the agent forms correct partial mentions, where partial mentions are considered correct if they occur in the training data a minimum number of times. The reader can find details about this in Section 4. The mention-based episodes terminate either when the agent labels the mention correctly or when it makes an error that cannot be corrected. We call these states terminal. Note that words that are not part of mentions are treated as single-word mentions with "O" labels.


\subsection{Learning}
As mention earlier, we approximate a state-action value function $Q(S,A)$ using a neural network $Q(S,A;\theta)$ that takes the representation of a state $S$ as input and outputs a value for each action $A \in \mathcal{A}$. We use Temporal Difference (TD) learning to learn $\theta$, the set of parameters used in the neural network. We specifically experimented with both SARSA and Q-learning TD algorithms (\citet{Sutton:1998:IRL:551283}).
The parameters $\theta$ are randomly initialized and the agent uses an $\epsilon$-greedy policy to explore. During exploration, we update the parameters $\theta$ by stochastic gradient descent (SGD) using gradients of the following loss function (Q-learning version):

$$
L(\theta) = (r+\gamma \displaystyle\max_{a'}Q(s',a';\theta^-) - Q(s,a,\theta))^2 
$$
where $r$ is the reward the agent observes after taking action $a$ from state $s$, arriving at state $s'$, and $\gamma$ is the reward discounting factor. The \textit{target} $Q(.;\theta^-)$ is a copy of $Q$ which is updated less frequently to stabilize the learning process as in \citet{schaul2016prioritized}. In order to keep the SGD update samples relatively independent, we compute the gradient of the loss function above for a batch of $K$ agents exploring different sentences (or mentions in the mention-based episodes). For each time step in the learning process, we randomly choose $K$ episodes from a pool of concurrent episodes $N$. When any of the concurrent episodes terminates, it is restarted by randomly selecting a new starting point from the training data. The algorithm for one-step Q-learning is given in Algorithm \ref{algorithm 1}.

\begin{algorithm}
\caption{Multi-agent one-step Q-learning algorithm for transition-based mention detection}\label{algorithm 1}
\begin{flushleft}
\item[] Input: $D$: training data, $K$: mini-batch size, $N$: number of concurrent learning episodes, $T$: total learning steps, $\alpha$: learning rate, $T_u$: $\theta^-$ update period,$\gamma$: reward discounting factor
\item[] Initialize $\theta$ for $Q(s,a;\theta)$ arbitrarily
\item[] Set $\theta^- \leftarrow \theta$
\item[] For i = 1 to N:
	\begin{itemize}
    \item[] Initialize $s_i$ by randomly choosing a starting point from $D$
    \end{itemize}
\item[] For t = 1 to T:
  \begin{itemize}
  \item[] B = Choose $K$ episodes from $N$ randomly
  \item[] For j = 1 to K:
    \begin{itemize}
    \item[] Set $i \leftarrow B[j]$
    \item[] $s_j \leftarrow s_i$
    \item[] Choose action $a_j$ from state $s_j$ according to $\epsilon$-greedy policy derived from $Q$
    \item[] Take action $a_j$ and observe state $s_j'$ and reward $r_j$
    \item[] If $s_j'$ is terminal:
      \begin{itemize}
        \item[] Set $R_j \leftarrow 0$
        \item[] Set $s_i$ to a starting point randomly chosen from $D$
      \end{itemize}
    \item[] Else:
      \begin{itemize}
        \item[] Compute $R_j \leftarrow \gamma \displaystyle\max_{a'}{Q(s_j',a',\theta^-)}$
        \item[] Advance episode $i$ by setting $s_i \leftarrow s_j'$
      \end{itemize}
    \end{itemize}
    \item[] Compute $\nabla\theta = \frac{1}{K}\displaystyle\sum_{j = 1}^{K}{\frac{\partial[r_j + R_j - Q(s_j,a_j,\theta)]^2}{\partial\theta}}$
    \item[] Update parameters $\theta \leftarrow \theta + \alpha \nabla\theta$
    \item[] If t mod $T_u$ == 0:
    \begin{itemize}
		\item[] Update $\theta^- \leftarrow \theta$
	\end{itemize}
  \end{itemize}
\end{flushleft}
\end{algorithm}

\section{Experiments}
We use two English and one Spanish mention detection data sets that vary in terms of size as well as the type system used (Table \ref{data}).
\begin{table}[h]
\caption{Data set statistics: number of mention types (excluding O) and number of tokens.}
  \label{data}
  \centering
  \begin{tabular}{l|llll}
    \toprule
    Data set & \#Labels & \#Train & \#Dev & \#Test  \\
    \midrule
    CoNLL En \cite{conll03} & 4 & 232K & 58K & 53K \\
	CoNLL  Spa \cite{conll03}& 4 & 270K & 54k & 52k \\
	OntoNotes En \cite{ontonotes}& 18 & 1778K & 271K & 186K \\
    \bottomrule
  \end{tabular}
\end{table}


\subsection{Models}
\label{exp-details}
\paragraph{Episodes and rewards} 
We initially set episodes to consist of individual sentences and we assign positive rewards only at the end of a sentence. Given a state-action pair $(s_i,a_i)$ at time step $i$, the reward is 0 if $s_{i+1}$ is not terminal, i.e. the buffer is not empty, otherwise the reward is the F-score or the accuracy (scaled to the [0:10] range) of the labels generated by the $a_{1:i}$ action sequence. 

Delaying the reward until the end of the sentence might strain the model unnecessarily as mention-level assignments can be evaluated independently. However, since the model can flexibly change the labels assigned as well as the extent/label of a past mention, it is not immediately clear how to reward actions before a terminal state is reached. To overcome this shortcoming, we train on mention-level episodes, where the boundaries are obtained from the gold annotation. Specifically, each mention and each non-mention word become individual episodes (of length 1 in the case of non-mentions). An episode is finished either when an unrecoverable error has been made or all the tokens of the mention have been processed. We assign a non-0 reward only when the episode is over, equal to $\pm 1$ scaled by the length of the mention. 

 
\paragraph{Learning}
We use one-step learning in the case of mention-based episodes and a larger TD step for sentence-based episodes. We update the parameters of the target network ($\theta^-$) every 5000 time steps. We use a budget of 3 million ($T$=3,000,000) for OntoNotes and 1 million for the much smaller CoNLL data sets. In both cases we use a batch size of 10 ($K$=10). 

\paragraph{Baselines}
We use as baseline a feed forward NN trained in a supervised fashion optimizing the word-level log-likelihood. We use the same network for all data sets, trained using SGD with a linearly decaying learning rate (0.5:0.001) over 15 epochs. 

\paragraph{NN-RL architecture}
Both the Q-function and the supervised baseline use the same following network architecture. The input is the concatenation of the target and context words (symmetric window of size 4) to which we add vectors for three standard features: dictionaries, capitalization flags and character-level representations (concatenation of the last hidden states of forward and backward LSTMs) as well as the previous 2 assigned labels. The word vectors are initialized with 300-dimensional pre-trained embeddings build on a concatenation of Gigaword, BOLT and Wikipedia (totaling 6 billion tokens). Both the additional feature vectors as well as the word vectors are fine-tuned during training (i.e. error is back-propagated to the input representation). We use one hidden layer of size 2000 and sigmoid as its activation function (relu for Q function model) and use 0.5 dropout on the two final layers.

We build a Q-function network by mapping the features used in this standard NN architecture to the new model. More precisely, features that are functions of the target word operate on the top word of the buffer, while the left context is represented as features of the stack. We use a max-pooling 2-gram convolution layer to obtain a fixed-length representation for the words grouped in one stack element.

\paragraph{Hyper-parameter tuning} 
Additional datasets (not used in this paper) have been used to determine robust settings for the NER/mention detection tasks, determining for example that SGD with decaying rate is better than Adagrad/Adadelta, 1 large hidden layer performs better than deeper architectures and modeling context as word concatenation outperformed the use of LSTMs. When available, hyperparameters reported in previous work as optimal (e.g. dimensionality of character-level LSTMs, dropout value) were employed.

\subsection{Results}

Table \ref{results1} reports F1 scores on the three data sets including 2016 state-of-the-art results, the supervised baseline system (SL) and reinforcement learning system (RL) with mention episodes. 

SARSA and Q-learning perform similarly. The best results are obtained with mention episodes and the RL models perform similarly to the supervised counterpart on CoNLL English, and drop up to 1 to 2 F points on OntoNotes and CoNLL Spanish respectively. As expected, sentence episodes (omitted in Table \ref{results1}) take much longer to converge and lead to overall poorer performance: Using a curriculum based on sentence length the models reach 87 F score on CoNLL English and only 56 on the more difficult Spanish data set, while using no curriculum halves the F scores. 



\begin{table}[h]
\caption{Mention detection F1 scores: State of the art, supervised baseline (SL) and Sarsa/Q-learning using mention-level episodes (RL). *-with tuned hyper-parameters.}
\label{results1}
\centering
\begin{tabular}{l|ll|ll|ll}
& \multicolumn{2}{c|}{CoNLL English} &\multicolumn{2}{c|}{OntoNotes} &\multicolumn{2}{c}{CoNLL Spanish}\\
\toprule
Model  & Dev&Test & Dev&Test & Dev&Test \\
\midrule
\citet{Lample16} & - & 90.94&-& &-&85.75\\
\citet{chiu-Nichols-16} & 94.03*& 91.62&84.57*&86.28&-&-\\
\midrule
SL &{93.06}& {89.85} & {86.39} & 85.59 & {83.82} & {84.64}\\
RL-Sarsa & 93.22 & 89.28& 85.15 & 85.35 & 81.29 & 83.61\\
RL-Qlearn & 93.25 & 89.17 & 85.23& 84.67 & 81.72 & 84.35 \\
\bottomrule
\end{tabular}
\end{table}

\paragraph{Sub-mentions} We further analyze the structure and the partial labels that the RL model assigns to investigate whether sub-mentions receive labels which are consistent with other occurrences as individual mentions. Specifically, for each  mention, we extract all its occurrences, if any, as a sub-sequence of a larger mention in the development and test data (for example Hong Kong and Hong Kong Disneyland). We further count how many times the original mention is identified as a chunk ([[Hong Kong] Disneyland] vs Hong [Kong Disneyland]]) as well as how many times the assigned label is consistent with the original label (e.g. [[Hong Kong]$_\mathrm{GPE}$ Disneyland] vs. [[Hong Kong]$_\mathrm{ORG}$ Disneyland]). 

We examine the previous RL model as well as an additional variant denoted (RL+partial rewards) which gives partial rewards before mention episodes are complete. More precisely, this model encourages "reasonable" sub-mention structures by rewarding partial decisions which lead to sub-mentions that have been seen in training. 

\begin{table}[h]
\caption{Labels assigned to sub-mentions broken-down by mention size. Chk: percentage of times the sub-mention is a chunk of the larger mention (Chk rnd: probability of random). Corr. lbl: Percentage of times the correct label is assigned to it. Chance: 0.08}
\label{partial}
\centering
\begin{tabular}{l|c|cc|cc|l}
& & \multicolumn{2}{c|}{RL} & \multicolumn{2}{c|}{RL+partial reward} &\\
Sz & Chk rnd. & Chk & Corr. lbl & Chk & Corr. lbl& Correct example \\
\midrule
1 & 1.0 & 1.0  & 0.11 & 1.0 & 0.65&[two [hours]$_{\mathrm{TIME}}$]$_{\mathrm{TIME}}$\\ 
2 & 0.44& 0.46 & 0.51 & 0.65 & 0.88 & [[New York]$_{\mathrm{GPE}}$ Stock Market]$_{\mathrm{ORG}}$ \\
3 & 0.33& 0.41 & 0.71& 0.60 & 0.93 & [the [Sino - Japanese]$_{\mathrm{NORP}}$ war]$_{\mathrm{EVENT}}$\\
4 & 0.31& 0.35 & 0.68& 0.62 & 0.92 & [end of [August of last year]$_{\mathrm{DATE}}$]$_{\mathrm{DATE}}$\\
\bottomrule
\end{tabular}
\end{table}

Table \ref{partial} shows the results broken down by (sub-)mention size. The mention words are grouped together only marginally above chance level. When words are grouped correctly, the labels assigned are correct more than half of the time. This accuracy is much lower for mentions of size 1 (11\%) where we observe the assigned label to be meaningless most of the times, e.g. [[China]$_{\mathrm{TIME}}$ Daily]$_{\mathrm{ORG}}$. We hypothesis this is happening because the model is using its look-ahead capabilities to assign correct labels and sees no benefit in making reasonable partial label assignments. The additional partial rewards used with the RL+partial model lead, as expected, to much more coherent sub-mention structures, both in terms of correct chunking and of labels assigned. Surprisingly though, this does not improve overall results, with RL+partial scoring $\approx$ 1F point bellow RL on the OntoNotes development dataset.


\paragraph{Label bias errors} We devise a similar method to identify errors which might be caused by label bias. More precisely, we extract cases such as New York Stock Market incorrectly tagged (correct label is ORG), possibly due to the [New York]'s strong GPE bias. The larger mention might be incorrectly identified in a number of ways. We hypothesize that label bias may play a role if either the entire mention or any prefix of it is tagged as GPE (Geographical and Political Entity), resulting in the extent being misclassified, such as in: [New York]$_{\mathrm{GPE}}$ [Stock Market]$_{\mathrm{ORG}}$. We are again using the development and test portions of OntoNotes and investigate the system output of the supervised and of the Q-learning transition-based models. The results are in Table \ref{bias}.


\begin{table}[h]
\caption{Number of potential label bias errors made by each model, broken down by size of inner mention (0 for mentions larger than 5). Examples contain assigned label as subscript and correct label as superscript.} 
\label{bias}
\centering
\begin{tabular}{l|ccl}
Men. size  & \#SL errors & \#RL errors & Example \\
\midrule
2 & 36 & 28 & [[Hong Kong] Disneyland]$_{\mathrm{GPE}}^{\mathrm{FACILITY}}$\\ 
3 & 15 & 5 & [[Bank of China] Tower]$_{\mathrm{ORG}}^{\mathrm{FACILITY}}$\\ 
4 & 2 & 0 & [[one hundred and twenty] dollar]$_{\mathrm{CARDINAL}}^{\mathrm{MONEY}}$\\ 
\bottomrule
\end{tabular}
\end{table}

As it can be observed, both models make very few mistakes of this type (the data contains approximately 30K mentions). We are however only extracting cases involving sub-mentions seen in the data and  might be underestimating the frequency of these errors. It may also be the case that given large annotated data of high quality, models make much fewer label bias errors. Investigating the several cases for which the RL corrects the supervised model, for the most cases, although not all, the sub-mention is correctly labeled, label which is further revised as the mention is processed. The word-level labels continue to be, as suggested by Table 4, mostly random, often rare labels such as LANGUAGE (LANG). Some examples include: [[Hong$_{\mathrm{LANG}}$ Kong$_{\mathrm{ORD}}$]$_{\mathrm{GPE}}$ [International$_{\mathrm{LANG}}$ Airport$_{\mathrm{LANG}}$]$_{\mathrm{FAC}}$]$_{\mathrm{FAC}}$, [[[one hundred]$_{\mathrm{CARD}}$ [and twenty]$_{\mathrm{ORG}}$]$_{\mathrm{CARD}}$ dollars]$_{\mathrm{MONEY}}$, [Asia [- [Pacific [Special [Olympic Games]$_{\mathrm{EVENT}}$]$_{\mathrm{EVENT}}$]$_{\mathrm{EVENT}}$]$_{\mathrm{EVENT}}$]$_{\mathrm{EVENT}}$ (omitting word labels)





\section{Conclusions and future work}
We have proposed a transition-based formulation for the mention detection task which generates  mention parses as labeled binary trees and allows the model to create sub-mentions and flexibly revise labeling decisions. We have used reinforcement learning to train this model, by rewarding complete mention labels consistent with gold-standard annotations. In this work we have used a convolutional layer to build compositional representations for phrases. In the future we plan to investigate some of the methods proposed in the vast existing literature on this topic, including ways to pretrain such composition functions using unsupervised, language-learning-type of objectives. 

We plan to also investigate ways to improve learning by bootstrapping with a policy learned in a supervised fashion, used to pretrain a critic in an actor-critic model. 

Finally, we want to investigate different ways of assigning rewards. In particular, we plan to use rewards as a means to integrate training signal that might not be available (or may be too expensive to compute) at decode time, such as rewarding partial mentions which are consistent with syntactic constituents, grouping together words which exhibit high co-occurrence statistics, or mentions that agree with other mention detection systems.



\small

\bibliography{bibl}
\bibliographystyle{plainnat}
\end{document}